# Creating synthetic night-time visible-light meteorological satellite images using the GAN method

Wencong Cheng[*1], Qihua Li[1], ZhiGang Wang[1], Wenjun Zhang[1], and Fang Huang[1]

[1]*Beijing Aviation Meteorological Institute, Beijing 100085, China*

## ABSTRACT

Meteorology satellite visible light images are critical for meteorologists. However, there is no satellite visible light channels data at nighttime, so we propose a method based on deep learning to create synthetic satellite visible light images during night. Specifically, to produce realistic-looking products, we trained a Generative Adversarial Networks (GAN) model. The model can generate satellite visible light images from corresponding satellite infrared channels data and numerical weather prediction(NWP) products. Considering explicitly evaluate the contributions of different satellite infrared channels and NWP products elements, we suggest using a channel-wise attention mechanic, e.g., SEBlock to quantitatively weigh the importance of different input data channels. The experiments based on the ERA5 reanalysis NWP products of ECMWF and the FY-4A meteorology satellite data show that the proposed method is effective to create realistic synthetic satellite visible light images during night.

**Article Highlights:**

● we propose a method based on deep learning (Specifically the GAN model), to create synthetic satellite visible light images during night.

- The proposed method can generate satellite visible light images from corresponding satellite infrared channels data and numerical weather prediction(NWP) products.

- we suggest using a channel-wise attention mechanic to quantitatively weigh the importance of different input data channels.

- The experiments based on the ERA5 reanalysis NWP products of ECMWF and the FY-4A meteorology satellite data show that the proposed method is effective to create realistic synthetic satellite visible light images during night.

## 1. Introduction

With the rapid development of satellite remote sensing technology, meteorology satellites play one of the most important roles in atmospheric analysis and forecast. Meteorology satellite images characterize the distribution of clouds, which can be used to track the evolution of large-scale weather systems. The data are used by meteorological researchers and the general public, providing continuous monitoring of clouds in certain regions of the Earth. With the help of meteorology satellites, researchers can make more accurate weather predictions. The meteorology satellite sensing data can be mapped to images in both visible light channels and infrared channels. There are multiple visible light sensing channels for most modern meteorology satellites, which can be merged to RGB true colors images. The merged true color visible light images are intuitive for users, that are rich in details and easy to be understood. Since the satellite visible light channels are not available during night, it is impossible to observe the clouds continuously by satellite visible light channels for 24 hours a day. Users need to switch the sensing channels and change the analysis mode when the day and the night shift, that is, using the visible light channels at daytime and the infrared channels at nighttime. For the above reasons, we investigate the problem of creating synthetic meteorology satellite visible light images during night.

Deep learning methods(LeCun *et al.*,2015) have been widely used in computer vision and other fields, which greatly improve the model capability of adapting data with complex spatial structures. The progress owes particularly to the introduction of convolutional neural networks (CNN) which characterize the probability distribution of the training data. But in the current researches, the deep generative methods based on CNN usually lead to generating blurred synthetic images by using the Euclidean distance(Pathak *et al.*,2016; Zhang *et al*.,2016; Mathieu *et al*.2016). Typically, a Euclidean distance(like L2 distance) is used as the loss function with the assumption that the data is drawn from a Gaussian distribution. This can be problematic in the case of multimodal distributions. Recent advances in the domain of image generation have been driven particularly by the invention of generative adversarial networks(GAN)(Goodfellow et al.,2014). GAN-based methods have achieved state-of-the-art results in producing very realistic images in an unsupervised setting(Radford *et al.*,2016; Lin *et al*.,2020; Zhang et al.,2017; Zhao *et al*.,2017). These methods use adversarial training processes to learn the model variables from the train data distribution. Conditional GAN(CGAN)(Isola *et al*.,2017) is a relatively straightforward variant of the basic GAN framework, which can learn the data distribution conditional to a given input.

The goal of this work is to create synthetic meteorology satellite visible channels images during night in a realistic way. We propose to learn the non-linear mapping from meteorology satellite infrared channels data and NWP products to meteorology satellite visible light channels images by introducing the CGAN model. We also use a channel-wise attention mechanic, e.g., Squeeze-and-Excitation Block(SEBlock)(Hu *et al*.,2018), at the front of the model to quantitatively weigh and evaluate the importance of the input data channels. The proposed method is evaluated on the ERA5 reanalysis data and Feng Yun-4A (FY-4A) geostationary meteorology satellite multi-channel data. Experimental results demonstrate that our method can effectively create realistic synthetic satellite visible light images during night.

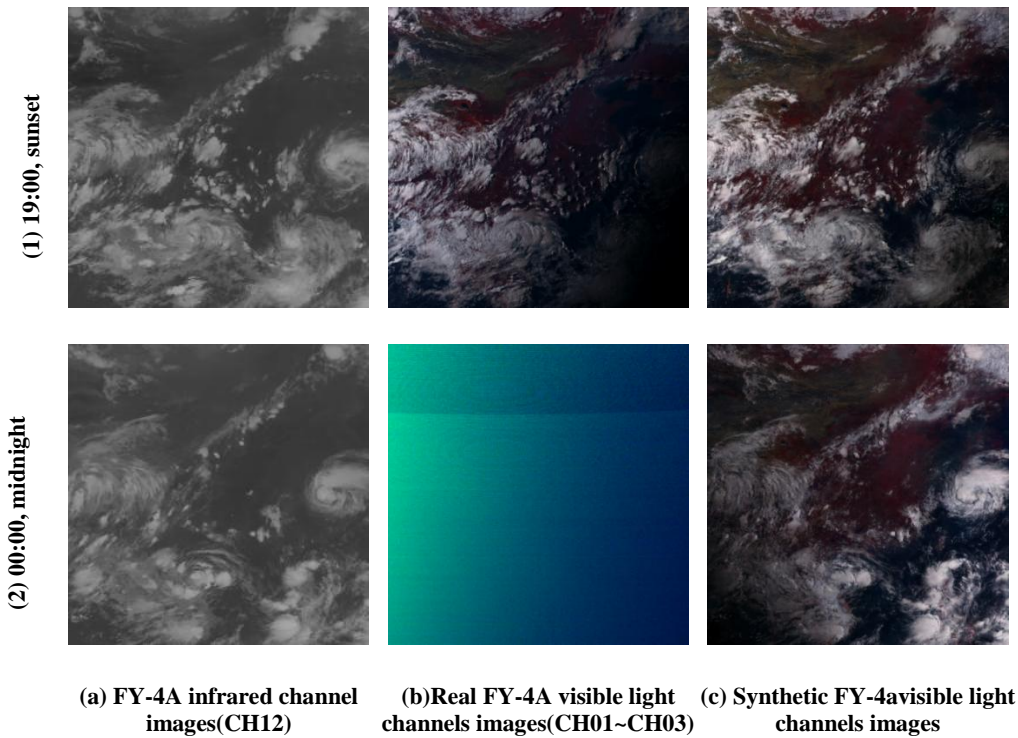

(a) FY-4A infrared channel images(CH12)  (b)Real FY-4A visible light channels images(CH01~CH03)  (c) Synthetic FY-4avisible light channels images

**Fig. 1.** Comparisons of satellite infrared channel images, visible light channels images, and synthetic visible light channels images. (**Row**: (1) images at 19:00, sunset. (2) images at 00:00, midnight. **Columns:** (a) FY-4A satellite infrared channel images (CH12). (b) real FY-4A satellite visible light channels images(CH01~CH03). (c) synthetic FY-4A satellite visible light channels images generated by this work.)

As shown in Figure 1, column (a) is a group of the images sensed by one of the FY-4A satellite infrared channels (CH12), which have data at daytime and nighttime. Column (b) is a group of the real FY-4A satellite visible light channels images. As the night approaches, there is no data at the bottom right part of the product firstly(shown in the row (1)), then there is totality blank but noise signals at the midnight(shown in row(2)). Column (c) is a group of the synthetic FY-4A satellite visible light channels images generated by our work. The results show that the proposed method in this paper can create realistic-looking synthetic satellite visible light channels images. To the best of our knowledge, this is the first application of using the GAN framework to create synthetic meteorology satellite visible light channels images during night.

## 2. Data

This work is conducted on the FY-4A meteorology satellite data and the ERA5 NWP (numerical weather prediction) reanalysis data. The training and testing data region cover the latitude from North $10°$ to $50°$ and longitude from East $110°$ to $150°$. The FY-4A satellite data can be downloaded from the website of the China National Satellite Meteorological Centre (NSMC) starting from April 2018. So we select the data ranging from April 1, 2018, to July 20, 2018, which are about 2500 hourly data samples to compose the training dataset. And the hourly data samples ranging from July 21 to July 22, 2018, are chosen to compose the test dataset.

(1) FY-4A meteorology satellite data

Feng Yun-4A (FY-4A) meteorology satellite, which is a second-generation geostationary meteorology satellite of China, was launched on December 11, 2016(Zhang *et al*., 2018; Zhang *et al*.,2019). It was fixed at the position of $99.5°$E above the equator.

The FY-4A satellite was equipped with four advanced optical instruments aboard, including an Advanced Geosynchronous Radiation Imager (AGRI), a Geostationary Interferometric Infrared Sounder (GIIRS), a Lightning Mapping Imager (LMI), and Solar X-EUV Imaging Telescope (SXEIT). The AGRI has 14 spectral bands from visible light to infrared (0.45-13.8μm) with high spatial (1 km for visible light channels, 2 km for near-infrared channels, and 4 km for remaining infrared channels) and temporal (full-disk sensing at the 15-min interval) resolutions. It is the most important instrument on the FY-4A satellite. In this work, we just use the AGRI data. The true color images (with RGB channels) can be merged by the AGRI visible light channels CH01, CH02, and CH03, which are used as the target products, the middlewave, and the longwave Infrared channels CH07~CH14 are used as parts of the model input data. The wavelengths, descriptions, and usages of the FY-4A satellite AGRI channels in this work are shown in Table 1.

**Table 1.** Wavelengths, descriptions, and usages of the FY-4A satellite AGRI channels in this work

| Channel ID | waveband (μm) | Description | Usage |
|:---:|:---:|:---:|:---:|
| 01 | 0.45~0.49 | Visible light channel corresponding to blue | Used as target data |
| 02 | 0.55~0.75 | Visible light channel corresponding to green | |
| 03 | 0.75~0.90 | Visible light channel corresponding to red | |
| 04 | 1.36~1.39 | Shortwave infrared channel, no data during night | not used in this work |
| 05 | 1.58~1.64 | Shortwave infrared channel, no data during night | |
| 06 | 2.1~2.35 | Shortwave infrared channel, no data during night | |
| 07 | 3.5~4.0(High) | middlewave and longwave infrared channels, having data during night | Used as parts of |
| 08 | 3.5~4.0(Low) | | |

| | | |
|---|---|---|
| 09 | 5.8～6.7 | model |
| 12 | 6.9～7.3 | input data |
| 11 | 8.0～9.0 | |
| 12 | 10.3～11.3 | |
| 13 | 11.5～12.5 | |
| 14 | 13.2～13.8 | |

(2)ERA5 NWP reanalysis data.

ERA5 is the 5th generation NWP reanalysis dataset from the European Centre for Medium-Range Weather Forecasts(ECMWF), which is one of the most widely used NWP reanalysis datasets (Hersbach *et al.*, 2020). This dataset contains meteorological records from 1950 to the near present. The most notable improvements of the ERA5 dataset from its predecessors are the finer spatial grid (31 km), the higher temporal resolution(hourly), the more number of vertical levels (137 levels), the richer assimilated sensing data sources, and a new NWP model (IFS Cycle 41r2). The ERA5 data can be downloaded from the website of the ECMWF. In this work, we use the hourly ERA5 NWP reanalysis products with the spatial resolution of 0.25 °×0.25 °as parts of the model input data. The ERA5 meteorological elements and the vertical pressure levels selected in this work are shown in Table 2. There are a total of 75 channels of the data considering NWP elements and levels.

**Table 2.** Elements and Levels of the ERA5 products used in this work

| elements | Unit | Levels |
|---|---|---|
| Fraction of cloud cover | (0 - 1) | |
| U-component of wind | m/s | 100/200/300/ |
| V-component of wind | m/s | 400/500/600/ |
| Vorticity | $s^{-1}$ | 700/800/850/ |
| Temperature | K | 900/950/1000 |
| Relative humidity | % | |

| Skin temperature | K | Single level |
|---|---|---|
| Total column cloud liquid water | kg m$^{-2}$ | Single level |
| Total column cloud ice water | kg m$^{-2}$ | Single level |

## 3. Method

The goal of this work is to use meteorology satellite infrared channels data and NWP products to generate corresponding satellite visible light channels images. Suppose we focus on a spatial region represented by an $M \times N$ grid in which the data consists of $M$ rows and $N$ columns. A satellite visible light channels image can be represented by a vector $S \in S^{M \times N}$, where $S$ denotes the domain of the satellite visible light channels data. $R_1$, $R_2$,……, $R_c$ is a set of the satellite infrared channels data, and $N_1$, $N_2$,……, $N_p$ is a set of NWP elements, $E$ denotes the conditional expectations. Then the synthetic satellite visible light image $\hat{S}$ is:

$$\hat{S} = E\big[S \mid R_1, R_2, \ldots\ldots, R_c, N_1, N_2, \ldots\ldots, N_p\big] \tag{1}$$

We train a Generative Adversarial Network (GAN) model to generate the satellite visible light images given the corresponding satellite infrared channels data and NWP products. GANs work by training two different networks: a generator network $G$ and a discriminator network $D$. $G$ generates the target samples as realistic as possible from the input data. $D$ has trained to estimate the probability of the input drawn from the real data, that is, $D$ tries to classify an input sample as 'real' or 'fake'(synthetic one). Following the GANs principle, both networks are trained simultaneously with $D$ trying to correctly discriminate between real and synthetic samples, while $G$ is trying to produce realistic samples that will confuse $D$.

For the generator network $G$, we use a U-Net(Ronneberger *et al.*,2015) neural network as the backbone structure, combined with a SEBlock as the front module. U-Net is an encoder-decoder module with skip-connections between mirrored layers in the encoder and decoder stacks, which is widely used in image segmentation and generation. Specifically, the U-Net module adds skip connections between each layer $i$ and layer $n$-$i$, where $n$ is the total number of layers. Each skip connection simply concatenates all channels at layer $i$ with those at layer $n$-$i$. There are 83 channels of the input data(8 for the FY-4A satellite infrared channels and 75 for the ERA5 NWP products). Since the resolution of the ERA5 NWP products is lower than the FY-4A satellite data, and an upsampling module is added to increase the ERA5 NWP products resolution to the satellite infrared channels data. A 'Squeeze and Extraction Block' (SEBlock) is introduced behind the input layers, to improve the representation capability by explicitly getting the channel-wise attention of the input data. The SEBlock module firstly performs a 'Squeeze' operation to squeeze the global spatial information of each channel into a channel descriptor, achieved by using global average pooling to generate channel-wise statistics. By the 'Squeeze' step we can get 83 real numbers corresponding to the 83 input data channels. Then, to make use of the information aggregated in the 'Squeeze' operation, an 'Excitation' operation is performed, which aims to capture the channel-wise attention by introducing a fully connected layer (FCLayer). The 'Excitation' operation maps the channel descriptors generated by the 'Squeeze' step to a set of channel weights and recalibrates the data by dotting product these channel weights. Discriminator $D$ is a standard classification convolution network. By estimating the probability of the module input being a 'real' satellite visible light image, $D$ can discern whether the module input data is a real satellite visible light image or a synthetic one. During the GAN training process, $D$ tries to correctly classify the real and the synthetic satellite visible light data, while *G* tries to generate the synthetic satellite visible light data as realistic as possible so that *D* cannot distinguish between them. To extract the mapping relationship between the input data *x* and the satellite visible light data *y*, the CGAN(Conditional GAN) model is used as the basic structure of the discriminator *D*, that is *x* and *y* are used together as the input of *D* (the discriminator *D* of the basic GAN only uses *y* as the input).

For *D*, we would like to find its parameters:

$$argmax_D log(D(x, y)) + log(1 - D(x, G(x, z)))$$ (2)

For *G*, we would like to optimize:

$$argmax_G log(D(x, G(x, z)))$$ (3)

The loss function for *D* is defined as:

$$L_D = L_{bce}(D(x, y), 1) + L_{bce}(D(x, G(x, z)), 0)$$ (4)

where：

$$L_{bce}(\hat{a}, a) = -\frac{1}{N} \sum_{i=1}^{N} (a_i log \hat{a}_i + (1 - a_i) log(1 - \hat{a}_i))$$ (5)

*N* is the number of samples in a minibatch of the model input, $a \in \{0, 1\}$ represents the label of the input data (1 for the real satellite visible light data and 0 for the synthetic data), and $\hat{a} \in [0, 1]$ is the label estimated by the discriminator *D*. When the value is close to 0, *D* assumes the input is the synthetic satellite visible light data, and when the value is close to 1, *D* assumes the input is the real satellite visible light data.

For *G*, as mentioned in Pathak *et al.*(2016) we use the loss composed by an adversarial term and a reconstruction L1 error. It is defined as:

$$L_G = \lambda_1 L_{bce}(D(x, G(x, z)), 1) + \lambda_2 |y - G(x, z)|$$ (6)

Where *y* is the corresponding ground truth satellite visible light data.

The generator *G* and discriminator *D* networks are described in detail in Figure 2. To train the GAN model, we alternatively train the generator network *G* with one batch of the input data and the discriminator *D* with two batches, in which one batch contains real samples and the other contains generated samples.

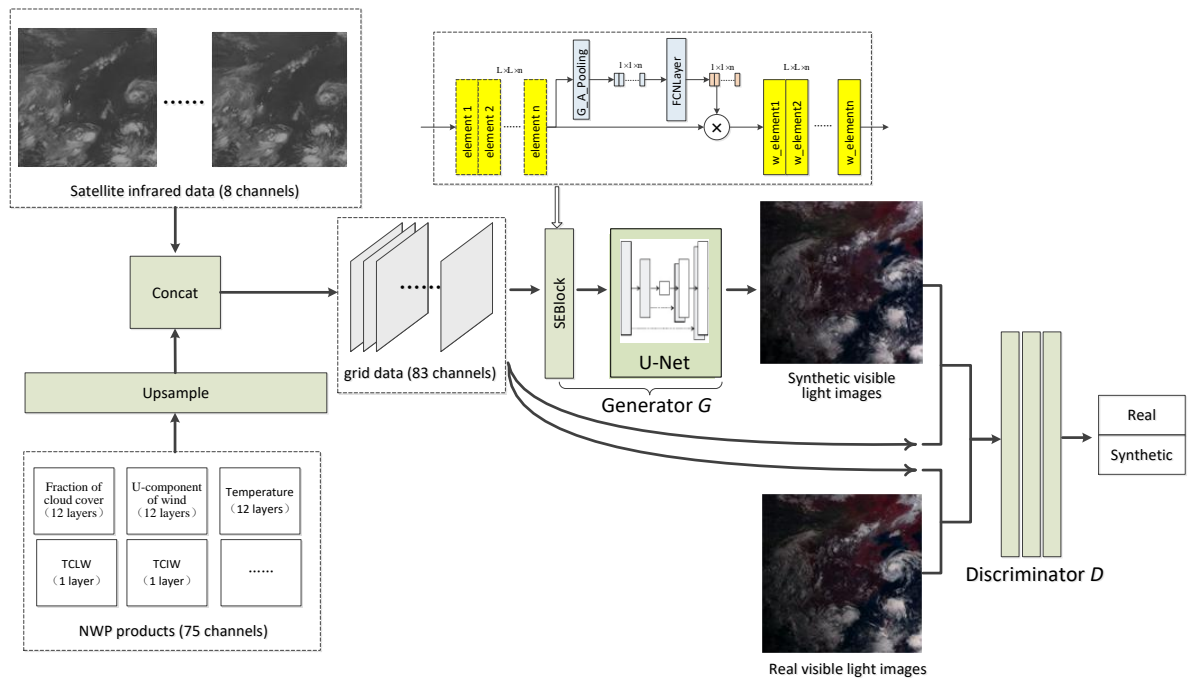

**Fig. 2.** The model architecture for creating synthetic satellite visible light images

## 4. Experiment Results and Analysis

To value the effect of the proposed method, We use mean absolute error (MAE) and root mean squared error (RMSE) as the quantitative metrics. In addition, the peak signal-to-noise ratio (PSNR) and structural similarity index measure (SSIM), which are commonly used in the domain of image reconstruction and generation are evaluated. For satellite visible light image with the resolution of $m{\times}n$, The formulas of the metrics used in this paper are shown as follow:

$$MAE = \frac{1}{mn}\sum_{i=0}^{m-1}\sum_{j=0}^{n-1}|I(i,j)-K(i,j)|$$

(7)

$$RMSE = \sqrt{\frac{1}{mn}\sum_{i=0}^{m-1}\sum_{j=0}^{n-1}[I(i,j)-K(i,j)]^2}$$

(8)

$$PSNR = 20{\times}log_{10}(\frac{MAX_I}{\sqrt{MSE}})$$

(9)

$$SSIM(x,y) = \frac{(2\mu_x\mu_y+c_1)(\sigma_{xy}+c_2)}{(\mu_x^2+\mu_y^2+c_1)(\sigma_x^2+\sigma_y^2+c_2)}$$

(10)

Where $MAX_I$ in (9) is the maximum grayscale (each channel) of the images, which is 255 in this work. In (10), $\mu_x$ and $\mu_y$ are the mean value of $x$ and $y$, $\sigma_x$ and $\sigma_y$ are the standard deviations of $x$ and $y$, $\sigma_{xy}$ is the covariance between $x$ and $y$, the positive constants $c_1$, $c_2$, and $c_3$ are used to avoid a null denominator.

The training data were collected from April 1 to July 20, 2018. The number of the training samples was 1057 (only using the hourly data from 08:00 to 17:00(daytime)). The resolution of the ERA5 NWP products is $160{\times}160$ as the input data, and the resolution of the satellite infrared channels products is $1000{\times}1000$ as the input data. The resolution of the targeting satellite visible light images is $1000{\times}1000$ as well.

We implemented the proposed model with Pytorch 1.0.1 deep learning framework and performed the training using Intel Xeon 4116$\times$2 CPU, 64GB RAM, and NVIDIA GTX 1080ti 11GB$\times$2 GPU. The Adam(Kingma *et al.*,2014) optimizer was used to train both the generator network and the discriminator network with a learning rate of 0.001, $\beta_1 = 0.5$, $\beta_2 = 0.999$. The mini-batch size is set to 8. We trained the model for 300 epochs, and the training time was approximately 25 hours.

Figure 3 displays selected examples of the synthetic satellite visible light images and corresponding real satellite images at different times, row (1) shows the real FY-4A satellite visible light true color images merged by visible light channels CH01, CH02, and CH03 as RGB channels, row (2) shows the synthetic satellite visible light images generated by this work, row (3) shows the synthetic satellite visible light images generated only by satellite infrared channels data, row (4) shows the synthetic satellite visible light images generated only by NWP products, row (5) shows one of the FY-4A satellite infrared channel (CH12, 10.3~11.3 μm) images and row (6) shows the total cloud amount element products of the ERA5.

Columns (a)~(e) demonstrate some cases at different times. Column (a) shows the samples at Beijing time 17:00 (that is 09:00 UTC, the following mentioned time are Beijing time as well) on July 21, 2018, before sunset. The real satellite visible light image had intact data. Column (b) shows the samples at 19:00 July 21, 2018, at sunset. The bottom right part of the real satellite visible light image became a blank space. Column (c) shows the samples at 0:00 July 22, 2018, at midnight. The real satellite visible light channels had no data at all. Column (d) shows the samples at 05:00 July 22, 2018, at sunrise. Most parts of the real satellite visible light image were blank, just a small top right part of the real satellite visible light image had data. Column (e) shows the samples at 8:00 July 22, 2018, when the sun had fully risen. The real satellite visible light image had intact data as Column (a).

Figure 3 shows that the proposed method can create realistic-looking satellite visible light images. The textures of the synthetic satellite visible light images are similar to the real samples at daytime. In Figure 3, there is a tropical cyclone in the middle right part of each image, which is the No.10 typhoon 'AMPIL' in 2018. Due to there being no visible light channels data during night, we cannot conduct the continuous observation of the tropical cyclone only using the satellite visible light images and need to switch to infrared channels during night. Taking the advantage of the proposed method, the continuous observation can be maintained in the satellite visible light channels, and the users needn't change the analysis mode to infrared channels during night.

According to the historical track data of tropical cyclones, 'AMPIL' had the strongest period from 09:00 July 21 to 03:00 July 22, and the maximum wind speed gradually reached 28m/s. Correspondingly in the synthetic satellite visible light images of Figure 3, we can find that the cyclone cloud system gradually shrank and became dense in columns (a) to (c). At 06:00 July 22, the maximum wind speed weakened to 25m/s, and at 09:00 it continually weakened to 23m/s. Correspondingly in the synthetic satellite visible light images of Figure 3, we can find the dense cloud system began to be loosened in column (d), and it became further loosened and more asymmetry in column (e). The cloud system shown in the synthetic visible light images can reflect the evolution of the typhoon exactly. We can see that the synthetic satellite visible light images provide more detailed information about 'AMPIL', which are helpful to analyze the characteristics and influence of the tropical cyclone. Haven the synthetic visible light images, we can observe and analyze complex meteorological events such as typhoons in detail during night.

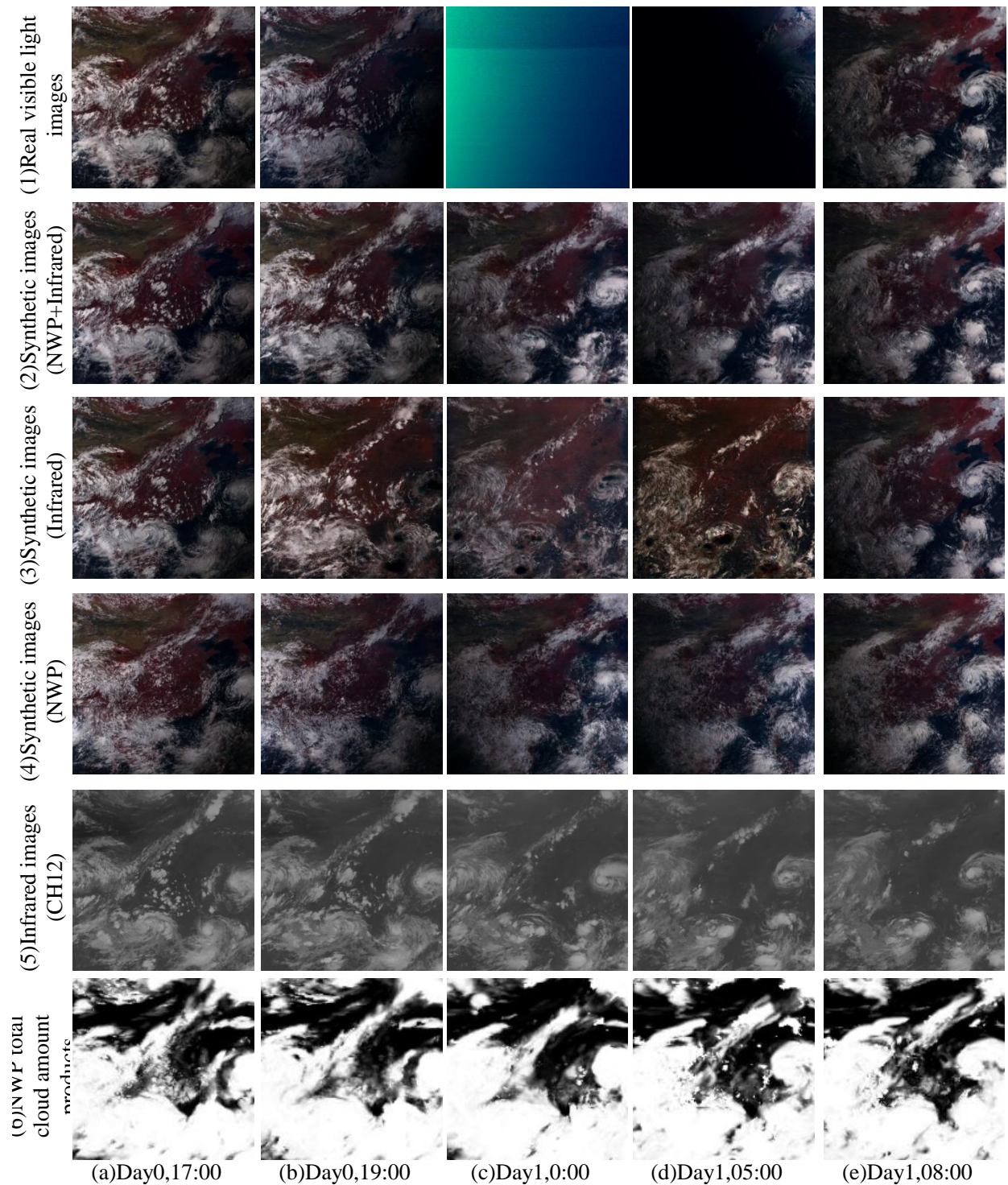

**Figure 3.** Samples of the synthetic satellite visible light images and the corresponding real satellite images, NWP products on July 21/22 2018. (**Rows:** (1) real satellite visible

light images. (2) synthetic satellite visible light images generated by NWP products + Infrared channels data. (3) synthetic satellite visible light images generated only by infrared channels data. (4) synthetic satellite visible light images generated only by NWP products. (5)satellite infrared channel images(CH12). (6) NWP total cloud amount element products. **Columns:** (a) Day0 (July 21, 2018), 17:00. (b)Day0, 19:00. (c)Day1 (July 22, 2018), 0:00. (d) Day1, 05:00. (e) Day1, 08:00.)

We can extract the weights in SEBlock of the model while generating the images. By comparing the weights corresponding to each input channel, we can roughly understand the contribution of each input channel to the final target product. The channel-wise attention weights in SEBlock at 00:00 July 22, 2018, are shown in Table 3. It can be found from the table that the infrared channels data have the greatest contribution to generating the synthetic visible light image during the night. The reason may be that the infrared channels data have the same resolutions as the visible light data, which is higher than the NWP products. So the infrared channels data can provide more detailed information. As for the NWP products, the contribution of the temperature element is highest among the NWP products according to the weights. We think the reason is the brightness temperature data sensed by satellite infrared channels change rapidly from day to night, and the NWP temperature element can be used as an important feature to adjust the effect of the generated target visible light images. We notice that the convolution operations in deep learning models have a certain adjustment ability, so the channel-wise attention weights extracted from the SEBlock module can only roughly reflect the contribution of each input data channel to the target products. It should not be considered as a quantitative metric to evaluate the importance of the input data channels.

**Table 3.** Channel-wise attention weights corresponding to the input elements at 00:00 July 22, 2018

| Categories | Elements | Weights |
|---|---|---|
| NWP products (multi-level elements) | Fraction of cloud cover | 0.11 |
| | U-component of wind | 0.18 |
| | V-component of wind | 0.19 |
| | Vorticity | 0.33 |
| | Temperature | 0.61 |
| | Relative humidity | 0.15 |
| NWP products (single levels products) | Skin temperature | 0.16 |
| | Total column cloud liquid water | 0.12 |
| | Total column cloud ice water | 0.05 |
| Satellite infrared channels data | CH07 | 0.81 |
| | CH08 | 0.6 |
| | CH09 | 0.25 |
| | CH10 | 0.76 |
| | CH11 | 1 |
| | CH12 | 1 |
| | CH13 | 0.97 |
| | CH14 | 0.49 |

Since there is no real satellite visible light channels data during night, to quantitatively evaluate the effect of the synthetic satellite visible light images during night, the optical flow method is introduced to extrapolate a forecast image at 18:00 from the visible light images at 16:00 and 17:00 on July 21, 2018. At 18:00 there are no real visible light channels data in the bottom right 1/4 part (corresponding to the area of latitude 40 °N~50 °N, longitude 20 °~30 °E) of the concerned region. So the bottom right 1/4 part of the forecast image extrapolated by the optical flow method at 18:00 is used as the test benchmark. The real, forecast, and synthetic visible light images are shown in Fig 4.

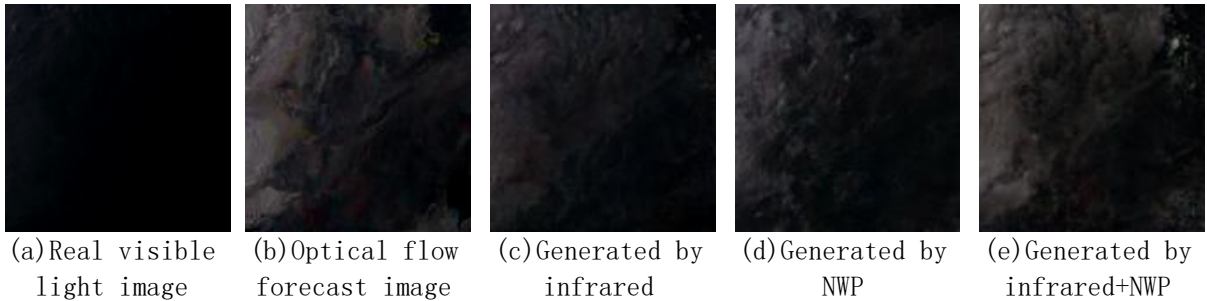

(a) Real visible light image   (b) Optical flow forecast image   (c) Generated by infrared   (d) Generated by NWP   (e) Generated by infrared+NWP

**Fig. 4.** Real, forecast, and synthetic visible light images.

Table 4 shows MAE (smaller values are better), RMSE (smaller values are better), PSNR (larger values are better), and SSIM (larger values are better) as the model input of infrared channels data, NWP products, and infrared channels data combined with NWP products. MAE and RMSE are evaluated using the albedo values of the visible light channels data, which range from 0 to 1.65. PSNR and SSIM are evaluated using the raw images directly. It can be found that the synthetic images generated by combined data quantitatively outperform the others, which shows the advantages of the proposed method.

**Table 4.** The quantitative evaluations according to the forecast visible light channels data

| Model input | Mean MAE | Mean RMSE | PSNR | SSIM |
|---|---|---|---|---|
| Infrared channels | 0.090 | 0.118 | 353.8 | 0.451 |
| NWP | 0.072 | 0.096 | 355.6 | 0.448 |
| Infrared channels+NWP | 0.062 | 0.082 | 357.0 | 0.480 |

The visual perceptions and the quantitative evaluations show that the proposed method can generate the synthetic satellite visible light channels images effectively.

## 5. conclusion and future work

In this work, we have presented a method using GAN architecture to create synthetic satellite visible light images during night. We leverage the idea from taking the combination of satellite infrared channels data and the NWP products as the model input. The experiment results in this paper suggest that compared to the models which input only with satellite infrared channels data and only with NWP products, the combination one learns to synthesize more realistic synthetic images. For future work, we will investigate merging the geographic information and other data to improve the effectiveness of the generated synthetic satellite visible light images, both in perceptual and quantitative similarities.